\RequirePackage{fix-cm}
\documentclass[twocolumn,final]{svjour3}

\smartqed
\usepackage{amsmath,amssymb}
\usepackage[utf8]{inputenc}
\usepackage[T1]{fontenc}
\usepackage[english]{babel}
\usepackage{csquotes}
\usepackage{microtype}
\usepackage{graphicx}
\usepackage{calc}
\usepackage[usenames,dvipsnames]{xcolor}
\usepackage{xspace}
\usepackage{listings}
\usepackage[ruled,linesnumbered]{algorithm2e}
\usepackage{nicefrac}
\usepackage{pgf}
\usepackage{pgfplots}
\usepackage{pifont}\usepackage{pgfplotstable}
\usepackage{tikz}
\usetikzlibrary{
    shapes, positioning, calc, backgrounds,
    shapes.arrows, arrows,
    arrows.meta,
    decorations.pathmorphing,
    decorations.markings,
}
\usepackage{url}
\usepackage[caption=false]{subfig}
\usepackage{caption}
\usepackage{scalefnt}
\usepackage[inline]{enumitem}
\usepackage[normalem]{ulem}
\usepackage{booktabs}
\usepackage{multirow}
\usepackage{setspace}
\usepackage{todonotes}
\usepackage[
  style=altlist,
  acronym=true
]{glossaries} \IfFileExists{zi4.sty}{\usepackage{zi4}
}{\IfFileExists{inconsolata.sty}{\usepackage{inconsolata}
}{}}
\usepackage[\ifx\debug\empty
    colorlinks
  \else
    hidelinks
  \fi]{hyperref}

\makeatletter
\def\cl@chapter{\@elt {theorem}}
\makeatother
\usepackage{cleveref}

\newlength\nspace
\newlength\fspace
\newlength\subfspace
\newlength\lstspaceAbove
\newlength\lstspaceBelow
\newlength\lstinlinespace
\newlength\spaceoffset
\newlength\spaceoffsetneg

\ifx\squeeze\empty
    \SetExpansion[shrink=50]{encoding={OT1,T1,TS1}}{}
    \SetAlgoSkip{negskip}
\else
\fi

\newcommand{\hipacc}{Hipacc\xspace}
\newcommand{\cpp}{\mbox{C\hspace{-.05em}\raisebox{.3ex}{\scalefont{0.65}{\textbf{++}}}}\xspace}

\newcommand{\OpenVX}{OpenVX\xspace}

\newcommand{\DomVX}{HipaccVX\xspace}

\newcommand{\ie}{\mbox{i.e.,}\xspace}
\newcommand{\eg}{\mbox{e.g.,}\xspace}

\xspaceaddexceptions{\,}

\crefname{equation}{eq.}{eqs.}
\Crefname{equation}{Eq.}{Eqs.}

\let\cref\Cref

}

\newcommand{\cmark}{\textcolor{ForestGreen}{\ding{51}}}\newcommand{\xmark}{\ding{55}} \graphicspath{{}}

\usepgfplotslibrary{units}
\pgfplotsset{compat=newest}
 \usepackage{calc}
\usepackage{listings}
\usepackage{xcolor}
\usepackage{scalefnt}

\makeatletter

\lstnewenvironment{lstinlinelisting}[1][]{\lstset{frame=none,numbers=none,aboveskip=\medskipamount,belowskip=\medskipamount,#1}}{}

\def\lst{\lstinline}

\makeatother

\lstdefinelanguage{OpenVX}{language=C++,
  morekeywords={vxCreateImage, vxCreateVirtualImage, vxReleaseContext,
                vxCreateGraph, vxVerifyGraph, vxProcessGraph, vxCreateContext,
                vxGaussian3x3Node, vxSobel, vxSobel3x3Node, vxMagnitudeNode, vxPhaseNode, vxAddNode,
                vxChannelExtractNode, vxAddParameterToKernel,vxCreateScalar,
                vxSetParameterByIndex, vxFinalizeKernel, vxCreateGenericNode,
                vxThresholdNode, vxCreateMatrix, vxCopyMatrix
                },
  keywords=[2]{vx_image, vx_context, vx_graph, VX_SUCCESS, VX_DF_IMAGE_VIRT, VX_DF_IMAGE_UYVY,
              VX_DF_IMAGE_S16, VX_DF_IMAGE_U8, VX_CHANNEL_Y, vx_matrix, vx_node
              },
  keywords=[3]{vxHipaccKernel, DOMVX_LOCAL},
keywordstyle=[2]{\color{Magenta}\bfseries},
  keywordstyle=[3]{\color{blue}\bfseries}
}

\lstdefinelanguage{HIPAcc}{language=C++,
  morekeywords={size_t, uchar, ushort, uint, uchar4, char4, ushort4, short4,
    uint4, int4, float4, double4},
  keywords=[2]{Image, Accessor, Kernel, IterationSpace, Mask, BoundaryCondition,
    Boundary, CLAMP, MIRROR, UNDEFINED, REPEAT, CONSTANT, input, output, convolve,
    iterate, reduce, Reduce, SUM, MIN, MAX, PROD, MEDIAN Domain},
keywordstyle=[2]{\color{blue}\bfseries}
}

\ifx\debug\empty

\makeatletter
\renewcommand{\todo}[2][]{\tikzexternaldisable \@todo[size=\footnotesize,caption={#2},#1]{\begin{spacing}{0.7}\fontfamily{phv}\fontseries{mc}\selectfont{\scalefont{.75}#2}\end{spacing}}\tikzexternalenable}
\makeatother

\let\MissingFigure\missingfigure
\renewcommand{\missingfigure}[1]{
  \tikzexternaldisable \MissingFigure{#1}
  \tikzexternalenable}

\else

\renewcommand{\todo}[2][]{}
\renewcommand{\missingfigure}[1]{}

\fi

\usepackage{suffix}

\newcommand{\ac}[1]{\gls{#1}}
\newcommand{\acf}[1]{\ifglsused{#1}{\acrfull{#1}}{\ac{#1}}}
\newcommand{\acs}[1]{\acrshort{#1}}
\newcommand{\acl}[1]{\acrlong{#1}}
\newcommand{\acp}[1]{\glspl{#1}}
\newcommand{\acfp}[1]{\ifglsused{#1}{\acrfullpl{#1}}{\acp{#1}}}
\newcommand{\acsp}[1]{\acrshortpl{#1}}
\newcommand{\aclp}[1]{\acrlongpl{#1}}
\newcommand{\acfi}[1]{\emph{\acrlong{#1}} (\textsc{\acrshort{#1}})}
\newcommand{\iac}[1]{a \ac{#1}}
\newcommand{\Iac}[1]{A \ac{#1}}
\WithSuffix\newcommand\ac*[1]{\ifglsused{#1}{\ac{#1}}{\ac{#1}\glsreset{#1}}}
\WithSuffix\newcommand\acf*[1]{\ifglsused{#1}{\acf{#1}}{\acf{#1}\glsreset{#1}}}
\WithSuffix\newcommand\acs*[1]{\ifglsused{#1}{\acs{#1}}{\acs{#1}\glsreset{#1}}}
\WithSuffix\newcommand\acl*[1]{\ifglsused{#1}{\acl{#1}}{\acl{#1}\glsreset{#1}}}
\WithSuffix\newcommand\acp*[1]{\ifglsused{#1}{\acp{#1}}{\acp{#1}\glsreset{#1}}}
\WithSuffix\newcommand\acfp*[1]{\ifglsused{#1}{\acfp{#1}}{\acfp{#1}\glsreset{#1}}}
\WithSuffix\newcommand\acsp*[1]{\ifglsused{#1}{\acsp{#1}}{\acsp{#1}\glsreset{#1}}}
\WithSuffix\newcommand\aclp*[1]{\ifglsused{#1}{\aclp{#1}}{\aclp{#1}\glsreset{#1}}}
\WithSuffix\newcommand\acfi*[1]{\ifglsused{#1}{\acfi{#1}}{\acfi{#1}\glsreset{#1}}}
\WithSuffix\newcommand\iac*[1]{a \ac*{#1}}
\WithSuffix\newcommand\Iac*[1]{A \ac*{#1}}
\newcommand{\newacro}[3][\empty]{\ifx#1\empty\newacronym{#2}{#2}{#3}\else\newacronym{#2}{#1}{#3}\fi}

\makenoidxglossaries

\pdfpageattr {/Group << /S /Transparency /I true /CS /DeviceRGB>>}
 \newacro{ALU}{Arithmetic Logic Unit}
\newacro{API}{Application Programming Interface}
\newacro{AST}{Abstract Syntax Tree}
\newacro{BGL}{Boost Graph Library}
\newacro{CV}{Computer Vision}
\newacro{CPU}{Central Processing Unit}
\newacro{DAG}{Directed Acyclic Graph}
\newacro{DSL}{Domain-Specific Language}
\newacro{DSP}{Digital Signal Processor}
\newacro{ECC}{Error-Correcting Code}
\newacro{FMA}{Fused Multiply-Add}
\newacro{FPGA}{Field Programmable Gate Array}
\newacro{GPU}{Graphics Processing Unit}
\newacro{HDR}{High Dynamic Range}
\newacro{HLS}{High-Level Synthesis}
\newacro{ISA}{Instruction Set Architecture}
\newacro{MPMD}{Multiple Program, Multiple Data}
\newacro{NPP}{NVIDIA Performance Primitive}
\newacro{OpenCV}{Open Source Computer Vision}
\newacro{PPnR}{Post Place and Route}
\newacro{MAD}{Multiply-Add}
\newacro{ROI}{Region of Interest}
\newacro{SPMD}{Single Program, Multiple Data}
\newacro{SDFG}{Synchronous Data Flow Graphs}
\newacro{SFU}{Special Function Unit}
\newacro{SSA}{Static Single Assignment}
\newacro{SPL}{Software Product Line}
\newacro[\hipacc{}]{HIPAcc}{Heterogeneous Image Processing Acceleration}

\newcommand{\CopyrightNotice}[2]{%
  \begin{picture}(0,0)(0,0)
    \put(#1){\parbox{\paperwidth-8em}{\sf \center {\footnotesize%
This is the author's version of the work.
Personal use of this material is permitted.
Permission from Springer must be obtained for all other uses, in any current or future media, including reprinting/republishing this material for advertising or promotional purposes,creating new collective works, for resale or redistribution to servers or lists, or reuse of any copyrighted component of this work in other works.
      }}}%
  \end{picture}
  \vspace{#2}
}

\def\makeheadbox{\relax}

\begin{document}

\title{Hipacc{VX}: Wedding of OpenVX and DSL-based Code Generation}

\author{
  M. Akif \"Ozkan \and
  Burak Ok \and
  Bo Qiao \and
  J\"urgen Teich \and
  Frank Hannig}


\institute{
  M. Akif \"Ozkan \and
  Burak Ok \and
  Bo Qiao \and
  J\"urgen Teich \and
  Frank Hannig \at
  Hardware/Software Co-Design, Department of Computer Science,
  Friedrich-Alexander University Erlangen-N\"urnberg (FAU), Germany\\
  \email{\{akif.oezkan, burak.ok, bo.qiao, teich, hannig\}@fau.de}
}


\date{}

\maketitle
\CopyrightNotice{-20,250}{0pt}

\begin{abstract}
Writing programs for heterogeneous platforms optimized for high performance is hard since this requires the code to be tuned at a low level with architecture-specific optimizations that are most times based on fundamentally differing programming paradigms and languages.
   \OpenVX promises to solve this issue for computer vision applications with a royalty-free industry standard that is based on a graph-execution model.
Yet, the \OpenVX' algorithm space is constrained to a small set of vision functions. This hinders accelerating computations that are not included in the standard.

   In this paper, we analyze \OpenVX vision functions to find an orthogonal set of computational abstractions. Based on these abstractions, we couple an existing Domain-Specific Language (DSL) back end to the \OpenVX environment and provide language constructs to the programmer for the definition of user-defined nodes. In this way, we enable optimizations that are not possible to detect with \OpenVX graph implementations using the standard computer vision functions.
   These optimizations can double the throughput on an Nvidia GTX GPU and decrease the resource usage of a Xilinx Zynq FPGA by 50\% for our benchmarks.
Finally, we show that our proposed compiler framework, called \DomVX, can achieve better results than the state-of-the-art approaches Nvidia VisionWorks and Halide-HLS.


\end{abstract}

\begin{table*}
  \caption{Available features in \OpenVX (VX), DSL compiler Hipacc (H), and our joint approach \DomVX (HVX).}
  \label{tab:allFeatures}
  \centering
  \scalebox{1}{
  \begin{tabular}{l|ccc}
    \toprule
    Features & VX & H & HVX \\
    \midrule
    Industrial standard (open, royalty-free)          & \cmark & \xmark & \cmark \\
    Community driven open-source implementations      & \xmark & \cmark & \cmark \\
    Well-known CV functions (\eg optical flow)
                                                      & \cmark & \xmark & \cmark \\
    High-level abstractions that adhere to distinct memory access patterns (\eg local)
                                                      & \xmark & \cmark & \cmark \\
    Custom node execution on accelerator devices (\ie OpenCL) & \cmark & \xmark & \cmark \\
    Acceleration of the custom nodes that are based on high-level abstractions
                                                      & \xmark & \cmark & \cmark \\
\toprule
  \end{tabular}
  }
\end{table*}

\section{Introduction}\label{sec:intro}

The emergence of cheap, low-power cameras and embedded platforms have boosted the use of smart systems with \ac{CV} capabilities in a broad spectrum of markets, ranging from consumer electronics, such as mobile, to real-time automotive applications and industrial automation, \eg semiconductors, pharmaceuticals, packaging.
The global machine vision market size was valued at \$16.0 billion already in 2018, and yet, is expected to reach a value of \$24.8 billion by 2023~\cite{bccResearch}.
A \ac{CV} application might be implemented on a great variety of hardware architectures ranging from \acp{GPU} to \acp{FPGA} depending on the domain and the associated constraints (\eg performance, power, energy, and cost).
Yet, for sophisticated real-life applications, the best trade-off is often achieved by heterogeneous systems incorporating different computing components that are specialized for particular tasks.

Optimizing \ac{CV} programs to achieve high performance on such heterogeneous systems usually goes along with sacrificing readability, portability, and modularity.
The programs need to be tuned at a low level with architecture-specific optimizations that are typically based on drastically different programming paradigms and languages (e.g., parallel programming of multicore processors using C++ combined with OpenMP; vector data types, libraries, or intrinsics to utilize the SIMD\footnote{Single Instruction, Multiple Data (SIMD) units are CPU components for vector processing, i.e., they execute the same operation on multiple data elements in parallel.} units of CPU; CUDA or OpenCL for programming GPU accelerators; hardware description languages such as Verilog or VHDL for targeting FPGAs).
Partitioning a program across different computing units, and accordingly, synchronizing the execution is difficult.
In order to achieve these ambitious goals, high development effort and architecture expert knowledge are required.

In 2014, the Khronos Group released \OpenVX as a C-based \acs{API} to facilitate cross-platform portability not only of the code but also of the performance for \ac{CV} applications~\cite{pressOpenVX10}. This is momentous since \OpenVX is the first (royalty-free) standard for a graph-based specification of \ac{CV} algorithms.
Yet, the \OpenVX' algorithm space is constrained to a relatively small set of vision functions. Users are allowed to instantiate additional code in the form of custom nodes, but these cannot be analyzed at the system-level by the graph-based optimizations applied from an \OpenVX back end.
Additionally, this requires users to optimize their implementations, who supposedly should not consider the optimizations of the performance.
Standard programming languages such as OpenCL do not offer performance portability across different computing platforms~\cite{steuwer2015generating,du2012cuda}.
Therefore, the user code, even optimized for one specific device, might not provide the expected high-performance when compiled for another target device.
These deficiencies are listed in \cref{tab:allFeatures}.

A solution to the problems mentioned above is offered by the community working on \acp{DSL} for image processing.
Recent works show that excellent results can be achieved when high-level image processing abstractions are specialized to a target device via modern metaprogramming, compiler, or code generation approaches~\cite{ragan2013halide,membarth2016hipacc,mullapudi2015polymage}.
These \acp{DSL} are able to generate code from a set of \emph{algorithmic abstractions} that lead to high-performance execution for diverse types of computing platforms.
However, existing \acp{DSL} lack formal verification, hence they do not ensure the safe execution of a user application whereas \OpenVX is an industrial standard.

In this paper, we couple the advantages of DSL-based code generation with \OpenVX (summarized in \cref{tab:allFeatures}).
We present a set of abstractions that are used as basic building blocks for expressing \OpenVX' standard \ac{CV} functions. These building blocks are suitable for generating optimized, device-specific code from the same functional description, and are systematically utilized for graph-based optimizations.
In this way, we achieve performance portability not only for \OpenVX' CV functions but also for user-defined kernels\footnote{A kernel in OpenVX is the abstract representation of a computer vision function~\cite{OpenVX13}.} that are expressed with these computational abstractions. The contributions of this paper are summarized as follows:
\vspace{0.5\spaceoffsetneg}
\begin{itemize}[leftmargin=\parindent,labelindent=0cm]\setlength{\itemsep}{0pt}
    \item We systematically categorize and specify \OpenVX' \ac{CV} functions by high-level abstractions that adhere to distinct memory access patterns (see \cref{sec:abstractions}).
\item We propose a framework called \DomVX, which is an \OpenVX implementation that achieves high performance for a wide variety of target platforms, namely, \acp{GPU}, CPUs, and \acp{FPGA} (see \cref{sec:implementation}). \item \DomVX supports the definition of custom nodes (\ie user-defined kernels) based on the proposed abstractions (see \cref{sec:dsl-backend}).
\item To the best of our knowledge, our approach is the first one that allows for graph-based optimizations that incorporate not only standard \OpenVX \ac{CV} nodes but also user-defined custom nodes (see \cref{sec:domvx-opt}), \ie optimizations across standard and custom nodes.
\end{itemize}

\begin{figure*}
    \centering
    \scalebox{0.8}{\includegraphics{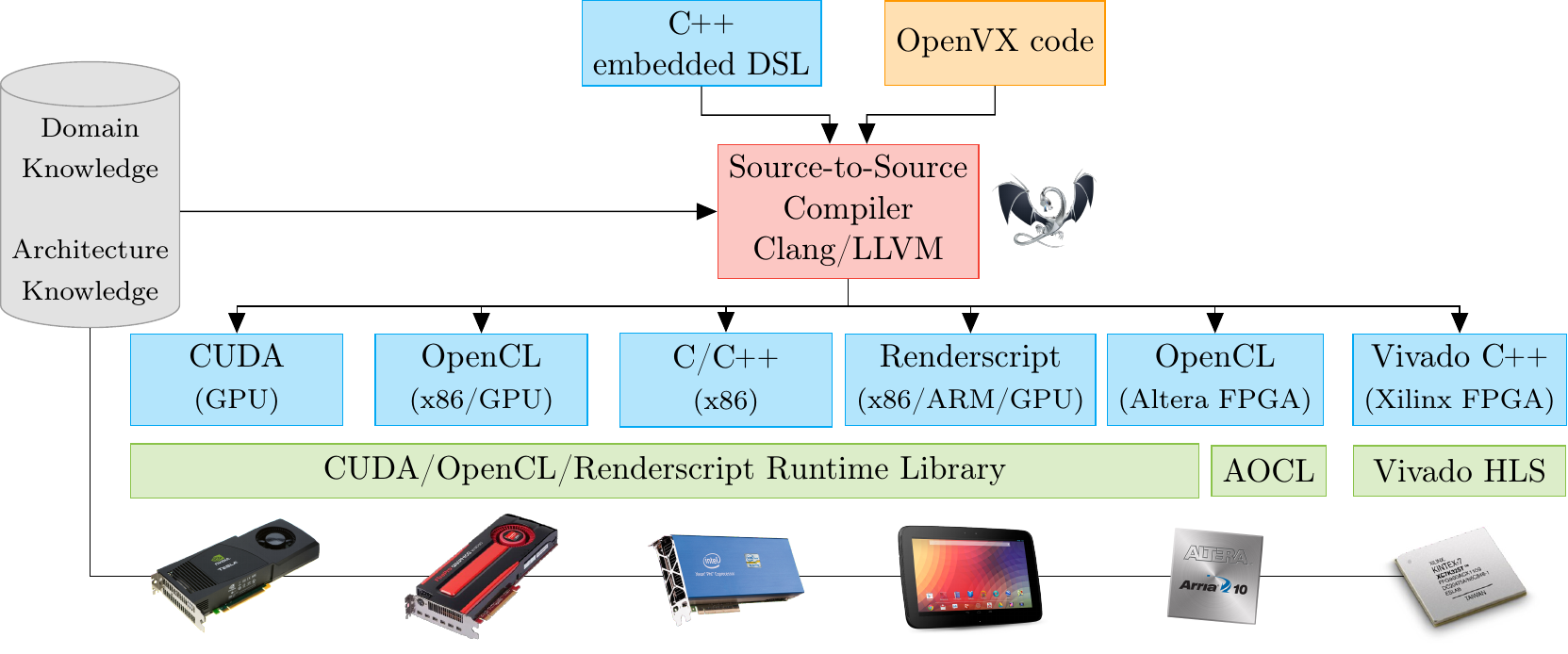}}
    \vspace{\fspace}
    \caption{\DomVX overview.}
    \label{fig:hipaccVX_overview}
    \vspace{\fspace}
\end{figure*}
\section{Related Work}\label{sec:related}

The \OpenVX specification is not constrained to a certain memory model as OpenCL and OpenMP, therefore enables better performance portability than traditional libraries such as OpenCV~\cite{rainey2014addressing}.
It has been implemented by a few major vendors, including Nvidia, Intel, AMD, and Synopsys~\cite{OpenVXresources}.
The authors of \cite{elliott2015supporting,yang2018making,mori2016design,zhang2018ds,tagliavini2016optimizing} focus on graph scheduling and design space exploration for heterogeneous systems consisting of \acp{GPU}, CPUs, and custom instruction-set architectures. Unlike the prior work, \cite{tagliavini2016enabling} suggests static \OpenVX compilation for low-power embedded systems instead of runtime-library implementations.
Our work is similar to this since we statically analyze a given \OpenVX application and combine the benefits of domain-specific code generation approaches~\cite{ragan2013halide,membarth2016hipacc,mullapudi2015polymage,hipaccIccad,pu2017programming,chugh2016dsl}.

Halide~\cite{ragan2013halide}, Hipacc~\cite{membarth2016hipacc}, and PolyMage~\cite{mullapudi2015polymage} are image processing \acp{DSL} that
provide language constructs and scheduling primitives to generate code that is optimized for the target device, \ie CPUs, \acp{GPU}.
Halide~\cite{ragan2013halide} decouples the algorithm description from scheduling primitives, \ie vectorization, tiling, while \hipacc~\cite{membarth2016hipacc} and PolyMage~\cite{mullapudi2015polymage} implicitly apply these optimizations on a graph-based description similar to \OpenVX.
CAPH~\cite{serot2013caph}, RIPL~\cite{stewart2016dataflow}, and Rigel~\cite{hegarty2016rigel} are image processing \acp{DSL} that generate optimized code for FPGAs.
Hipacc-FPGA~\cite{hipaccIccad} supports HLS tools of both Xilinx and Intel,
while Halide-HLS~\cite{pu2017programming}, PolyMage-HLS~\cite{chugh2016dsl}, and RIPL only target Xilinx devices.
CAPH relies upon the actor/dataflow model of computation to generate VHDL or SystemC code.
Our approach could also be used to implement OpenVX by these image processing DSLs.

There is no publicly available \OpenVX implementation for Xilinx \acp{FPGA} to the best of our knowledge.
Intel OpenVino~\cite{OpenVINO} provides a few example applications that are specific to Arria-10 \acp{FPGA}.
Taheri et al.~\cite{taheri2018acceleration} provide some initial results for \acp{FPGA}, where the main attention is the scheduling of statistical kernels (\ie histogram).
The image processing DSLs in~\cite{hipaccIccad,chugh2016dsl} use similar techniques to implement user applications as a streaming pipeline.
\Cref{sec:dataflow} shows how to instrument these techniques for the \OpenVX API.
Omidian et al.~\cite{omidian2018janus} present a heuristic algorithm for the design space exploration of \OpenVX graphs for \acp{FPGA}.
This algorithm could be simplified by using \DomVX' abstractions (see \cref{sec:abstractions}) instead of \OpenVX' \ac{CV} functions.
Then it could be used in conjunction with \DomVX to explore the design space of hardware/software platforms.
Moreover, Omidian et al.~\cite{omidian2018overlay} suggest an overlay architecture for \ac{FPGA} implementations of \OpenVX.
The proposed overlay implementation requires the optimized implementation of \OpenVX' CV functions, which could be generated by \DomVX.
Furthermore, an overlay architecture based on \DomVX's abstractions, which is a smaller set of functions compared to \OpenVX CV functions, could reduce resource usage in \cite{omidian2018overlay}.

Intel's \OpenVX implementation~\cite{ashbaugh2017opencl} is the first work extending the OpenVX standard with an interoperability API for OpenCL.
This is supported in \OpenVX v1.3~\cite{OpenVX13}.
Yet, performance portability still cannot be assured for the custom nodes.
An OpenCL code tuned for a specific CPU might perform very poorly on \acp{FPGA} and \ac{GPU} architectures~\cite{steuwer2015generating,du2012cuda}.
Contrarily to our approach, the performance of this approach relies on the user code.

 \section{\OpenVX and Image Processing DSLs}

In the following \cref{sec:openvx,,sec:hipacc}, we briefly explain the programming models of \OpenVX and image processing DSLs, respectively.
Then, we discuss the complementary features of these approaches in \cref{sec:combination}, which are the motivation of this work.

\subsection{\OpenVX programming model}\label{sec:openvx}

\begin{figure*}[!t]\centering
\includegraphics[width=0.9\linewidth]{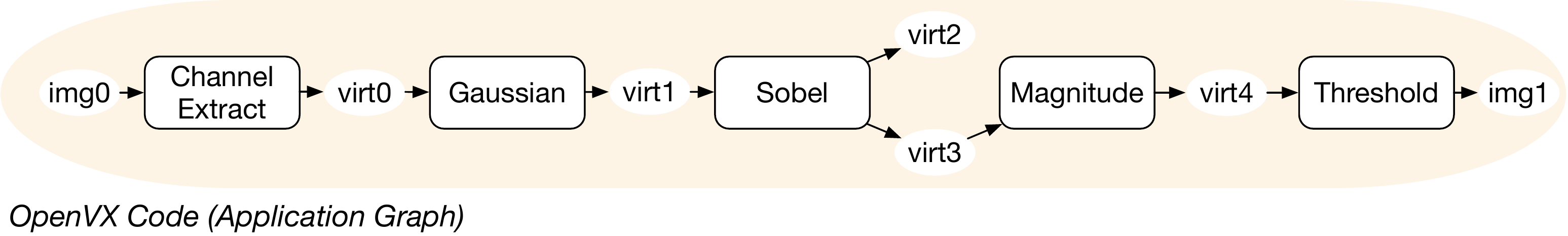}\vspace{\nspace}
\captionsetup[figure]{width=0.91\linewidth}
\captionof{figure}{
Graph representation for the \OpenVX code given in \cref{lst:openvx-ex}.
The output image (\emph{img1}) contains solely the horizontal edges extracted from the input image (\emph{img0}).
The \emph{virt2} image is defined only because OpenVX' Sobel function returns both horizontal and vertical edges.
This redundant computation is eliminated during the optimization passes of our \DomVX compiler framework (see \cref{sec:domvx-opt-comp}).
 \label{fig:teaser-appgraph}}
\vspace{\fspace}
\end{figure*}

\OpenVX is an open, royalty-free C-based standard for the cross-platform acceleration of computer vision applications.
The specification does not mandate any optimizations or requirements on device execution; instead, it concentrates on software abstractions that are freed from low-level platform-specific declarations.
The \OpenVX \acs{API} is totally opaque; that is, the memory hierarchy and device synchronization are hidden from the user.
Typically, platform experts of the individual hardware vendors provide optimized implementations of the \OpenVX \acs{API}~\cite{OpenVXresources}.

\Cref{lst:openvx-ex} shows an example \OpenVX code for a simple edge detection algorithm, for which the application graph is shown in \cref{fig:teaser-appgraph}.
An application is described as a \ac{DAG}, where nodes represent \ac{CV} functions (see \Crefrange{ll:vx-node1}{ll:vx-node5}) and data objects, \ie images, scalars (see \Crefrange{ll:vx-img0}{ll:vx-virt4}), while edges show the dependencies between nodes.
All \OpenVX \emph{objects} (\ie \emph{graph}, \emph{node}, \emph{image}) exist within a \emph{context} (\cref{ll:vx-context}).
A \emph{context} keeps track of the allocated memory resources and promotes implicit freeing mechanisms at \lst[language=OpenVX]{release} calls (\cref{ll:vx-release}).
A \emph{graph} (\cref{ll:vx-graph}) solely operates on the data objects attached to the same context.

The data objects that are used only for the intermediate steps of a calculation, which can be inaccessible for the rest of the application, should be specified as \emph{virtual} by the users.
Virtual data objects (\ie \emph{virtual images} defined in \Crefrange{ll:vx-virt1}{ll:vx-virt4}) cannot be accessed via read/write operations.
This paves the way for system-level optimizations applied in a platform-specific back end, \ie host-device data transfers or memory allocations~\cite{rainey2014addressing}.

The execution is not eager; an \OpenVX graph must be \emph{verified} (\cref{ll:vx-verify}) before it is \emph{executed} (\cref{ll:vx-proc}). The verification ensures the safe execution of the graph and resolves the implementation types of virtual data objects.
The \OpenVX standard mandates that a verification procedure must
\begin{enumerate*} [label=(\itshape\roman*\upshape)]
\item validate the node parameters (\ie presence, directions, data types, range checks), and
\item assure the graph connectivity (detection of cycles),
\end{enumerate*}
at the minimum~\cite{OpenVX12}.
Optimizations of an \OpenVX back end should be performed during the verification phase.
The verification is considered to be an initialization procedure and might restructure the application graph before the execution.
A verified graph can be executed repeatedly for different input parameters (\ie a new frame in video processing).

\begin{lstlisting}[
    float =!t,
    basicstyle=\scalefont{.65}\tt,
    language=OpenVX,
    caption={OpenVX code for an edge detection algorithm. The application graph derived for this OpenVX program is shown in \cref{fig:teaser-appgraph}.},
    aboveskip=\lstspaceAbove,
    belowskip=\lstspaceBelow,
    label ={lst:openvx-ex},
    ]
vx_context context = vxCreateContext(); $\label{ll:vx-context}$
vx_graph graph = vxCreateGraph(context);$\label{ll:vx-graph}$

vx_image img[] = {$\label{ll:vx-img0}$
  vxCreateImage(context, width, height, VX_DF_IMAGE_UYVY),$\label{ll:vx-img1}$
  vxCreateImage(context, width, height, VX_DF_IMAGE_U8)}; $\label{ll:vx-img2}$

vx_image virt[] = {
  vxCreateVirtualImage(graph, 0, 0, VX_DF_IMAGE_VIRT), $\label{ll:vx-virt1}$
  vxCreateVirtualImage(graph, 0, 0, VX_DF_IMAGE_VIRT), $\label{ll:vx-virt2}$
  vxCreateVirtualImage(graph, 0, 0, VX_DF_IMAGE_VIRT), $\label{ll:vx-virt3}$
  vxCreateVirtualImage(graph, 0, 0, VX_DF_IMAGE_VIRT)};$\label{ll:vx-virt4}$

vxChannelExtractNode(graph, img[0], VX_CHANNEL_Y, virt[0]);$\label{ll:vx-node1}$
vxGaussian3x3Node(graph, virt[0], virt[1]);                $\label{ll:vx-node2}$
vxSobel3x3Node(graph, virt[1], virt[2], virt[3]);          $\label{ll:vx-node3}$
vxMagnitudeNode(graph, virt[3], virt[3], virt[4]);         $\label{ll:vx-node4}$
vxThresholdNode(graph, virt[4], thresh, img[1]);           $\label{ll:vx-node5}$

status = vxVerifyGraph(graph); $\label{ll:vx-verify}$
if (status == VX_SUCCESS)
  status = vxProcessGraph(graph); $\label{ll:vx-proc}$

vxReleaseContext(&context); $\label{ll:vx-release}$
\end{lstlisting}

\subsubsection{Deficiencies of \OpenVX}\label{sec:prob-openvx}

As mentioned above, the \OpenVX standard relieves an application programmer from low-level, implementation-specific descriptions, and thus enables portability across a variety of computing platforms.
In \OpenVX, the smallest component to express a computation is a graph node (\eg \lst[language=OpenVX]{vxGaussian3x3Node}) from the set of base \ac{CV} functions.
However, these \ac{CV} functions are restricted to a small set since \OpenVX has a tight focus on cross-platform acceleration~\cite{OpenVX13}.
Custom nodes can be added to extend this functionality\footnote{The support for the execution of a user code (custom node) as part of an application graph on an accelerator device was introduced only recently (August 2019) with the release of \OpenVX v1.3~\cite{OpenVX13}.
Previous versions~\cite{OpenVX12} constraint the usage of the user-defined kernels to the host platform and required them to be implemented as \cpp kernels.
},
but, they leave the following issues unresolved:
\begin{enumerate*} [label=(\itshape\roman*\upshape)]
\item Users are responsible for the performance of a custom node, who supposedly should not consider performance optimizations.
\item Portability of performance cannot be enabled for the cross-platform acceleration of user code.
\item The graph optimization routines cannot analyze custom nodes.
\end{enumerate*}

\begin{figure}
    \centering
    \includegraphics[width=0.8\linewidth]{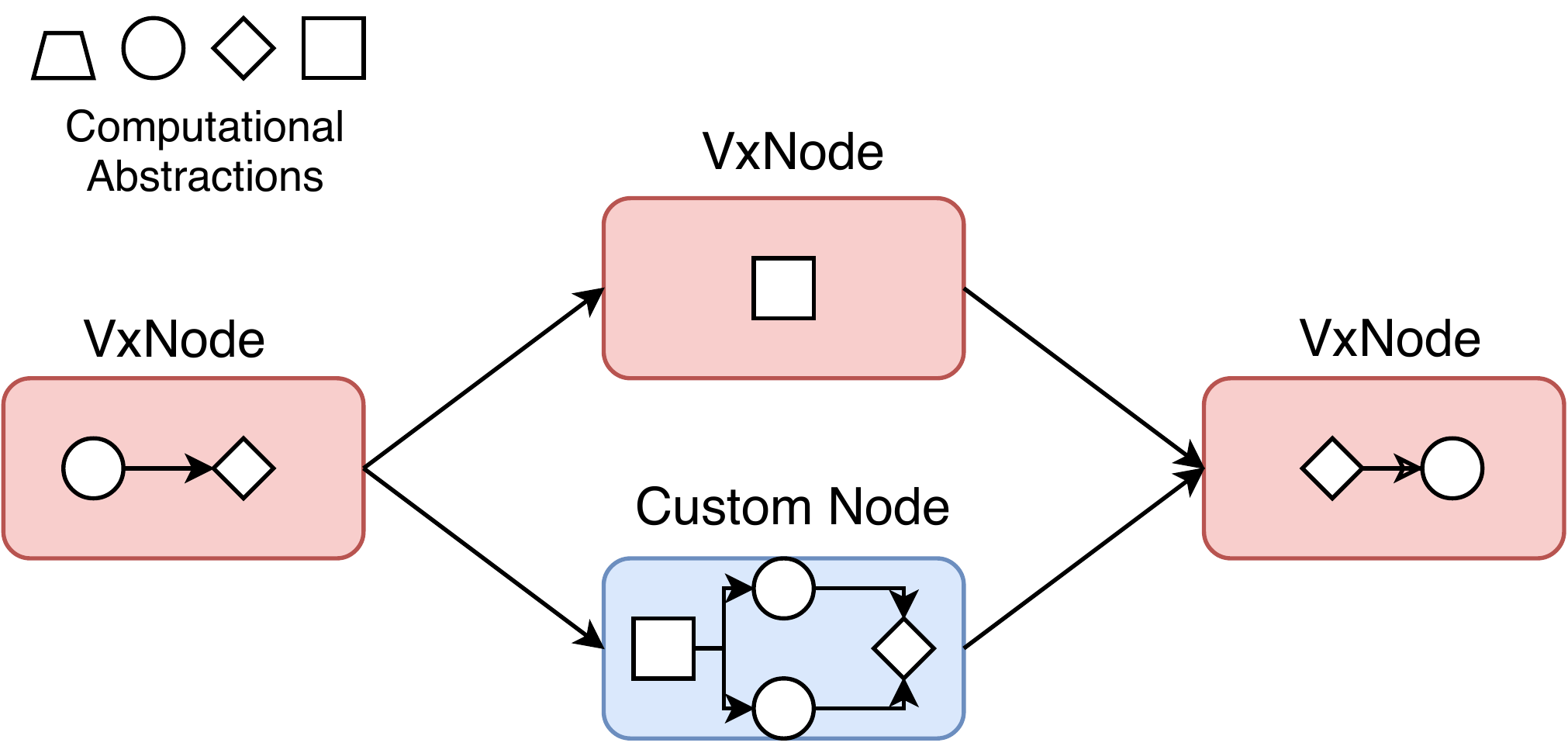}\vspace{\nspace}
    \captionof{figure}{\DomVX enables performance portability for user-defined code by representing \OpenVX' \ac{CV} functions as well as custom nodes by a small set of computational abstractions.}
    \label{fig:customNode}
    \vspace{\fspace}
\end{figure}

For instance, consider \cref{fig:customNode} that depicts an \OpenVX application graph with three \ac{CV} function nodes (red) and a user-defined kernel node (blue).
A \ac{GPU} back end would offer optimized implementations of the vxNodes (\eg Gauss), but the user code (custom node) is a black box for the graph optimizations.

Programming models such as OpenCL can be used to implement custom nodes.
This enables functional portability across a great variety of computing platforms.
However, the user should have \emph{expertise} in the target architecture in order to optimize an implementation for high performance.
Furthermore, OpenCL cannot assure the portability of the performance since the code needs to be tuned according to the target device, \ie usage of device-specific synchronization primitives, exploitation of texture memory if available, usage of vector operations, or different numbers of hardware threads~\cite{steuwer2015generating,du2012cuda}.
In fact, an OpenCL code optimized for an \ac{ISA} has to be ultimately rewritten for an \ac{FPGA} implementation in order to deliver high-performance~\cite{ozkan2016fpga}.

\subsection{Image Processing DSLs}\label{sec:hipacc}

Recently proposed \ac{DSL} compilers for image processing, such as Halide~\cite{ragan2013halide}, Hipacc~\cite{membarth2016hipacc}, and PolyMage~\cite{mullapudi2015polymage}, enable the portability of high-performance across varying computing platforms.
All of them take as input a high-level, functional description of the algorithm and generate platform-specific code tuned for the target device.
In this work, we use \hipacc to present our approach.

\hipacc provides language constructs that are embedded into \cpp for the concise description of computations. Applications are defined in a \ac{SPMD} context, similar to kernels in CUDA and OpenCL.
For instance, \cref{lst:hipacc-host} shows the description of a discrete Gaussian blur filter application.
First, a \lst[language=HIPAcc]{Mask} is defined in \cref{ll:hipacc-host-mask} from a constant array.
Then, input and output \lst[language=HIPAcc]{Image}s are defined as \cpp objects in \cref{ll:hipacc-host-im-in,ll:hipacc-host-im-out}, respectively.
\emph{Clamping} is selected as the image boundary handling mode for the input image in \cref{ll:hipacc-host-bh}.
The whole input and output images are defined as \ac{ROI} by the \lst[language=HIPAcc]{Accessor} and \lst[language=HIPAcc]{IterationSpace} objects that are specified in \cref{ll:hipacc-host-acc,ll:hipacc-host-is}, respectively.
Finally, the Gaussian kernel is instantiated in \cref{ll:hipacc-host-inst} and executed in  \cref{ll:hipacc-host-exec}.

\cref{lst:hipacc-conv} describes the actual \emph{operator kernel} for the Gaussian shown in \cref{lst:hipacc-host}.
The \lst[language=HIPAcc]{LinearFilter} is a user-defined class that is derived from \hipacc's \lst[language=HIPAcc]{Kernel} class, where the \lst[language=HIPAcc]{kernel} method is overridden.
There, a user describes a convolution as a lambda function using the \lst[language=HIPAcc]{convolve()} construct, which computes an output pixel (\lst[language=HIPAcc]{output()}) from an input window \mbox{(\lst[language=HIPAcc]{input(mask)})}.
\hipacc's compiler utilizes Clang's \ac{AST} to specialize the lambda function according to the selected platform and generates device-specific code that provides high-performance implementations when compiled with the target architecture compiler.
We refer to~\cite{membarth2016hipacc,hipaccIccad} for more detailed explanations, further programming language constructs of \hipacc as well as corresponding code generation techniques.

\begin{lstlisting}[
    float =!t,
    basicstyle=\scalefont{0.64}\tt,
    language=HIPAcc,
    frame=tb,
    label ={lst:hipacc-host},
    belowskip=0.5\lstspaceBelow,
    caption={\hipacc application code for a Gaussian filter. It instantiates the $\textrm{\lst[basicstyle=\ttfamily]{LinearFilter}}$ Kernel given in \cref{lst:hipacc-conv}.}
    ]
// filter mask for Gaussian blur filter
const float filter_mask[3][3] = {
  { 0.057118f, 0.124758f, 0.057118f },
  { 0.124758f, 0.272496f, 0.124758f },
  { 0.057118f, 0.124758f, 0.057118f }
};
Mask<float> mask(filter_mask);                           $\label{ll:hipacc-host-mask}$

// input and output images
size_t width, height;
uchar *image = read_image(&width, &height, "input.pgm");
Image<uchar> in(width, height, image);                   $\label{ll:hipacc-host-im-in}$
Image<uchar> out(width, height);                         $\label{ll:hipacc-host-im-out}$

// reading from in with clamping as boundary condition
BoundaryCondition<uchar> cond(in, mask, Boundary::CLAMP);$\label{ll:hipacc-host-bh}$
Accessor<uchar> acc(cond);                               $\label{ll:hipacc-host-acc}$

// output image (region of interest is the whole image)
IterationSpace<uchar> iter(out);                         $\label{ll:hipacc-host-is}$

// instantiate and launch the Gaussian blur filter
LinearFilter Gaussian(iter, acc, mask, 3);$\label{ll:hipacc-host-inst}$
Gaussian.execute();                       $\label{ll:hipacc-host-exec}$
\end{lstlisting}
\begin{lstlisting}[
    float =!t,
    basicstyle=\scalefont{0.64}\tt,
    language=HIPAcc,
    frame=tb,
    label ={lst:hipacc-conv},
    belowskip=0.5\lstspaceBelow,
    caption={\hipacc kernel code for an FIR filter.}
    ]
class LinearFilter: public Kernel <uchar > {
  // ...
  public:
    LinearFilter(Accessor<uchar> &input, // input image
             IterationSpace<uchar> &out, // output image
             Mask<float> &mask)          // mask
             : {/* ... */}

  void kernel() { // convolve -> local operator
    output() = convolve(mask, Reduce::SUM, [&] () -> uchar {
          return mask() * input(mask);
    });
  }
};
\end{lstlisting}

\begin{figure*}[!t]\centering
  \includegraphics[width=0.85\linewidth]{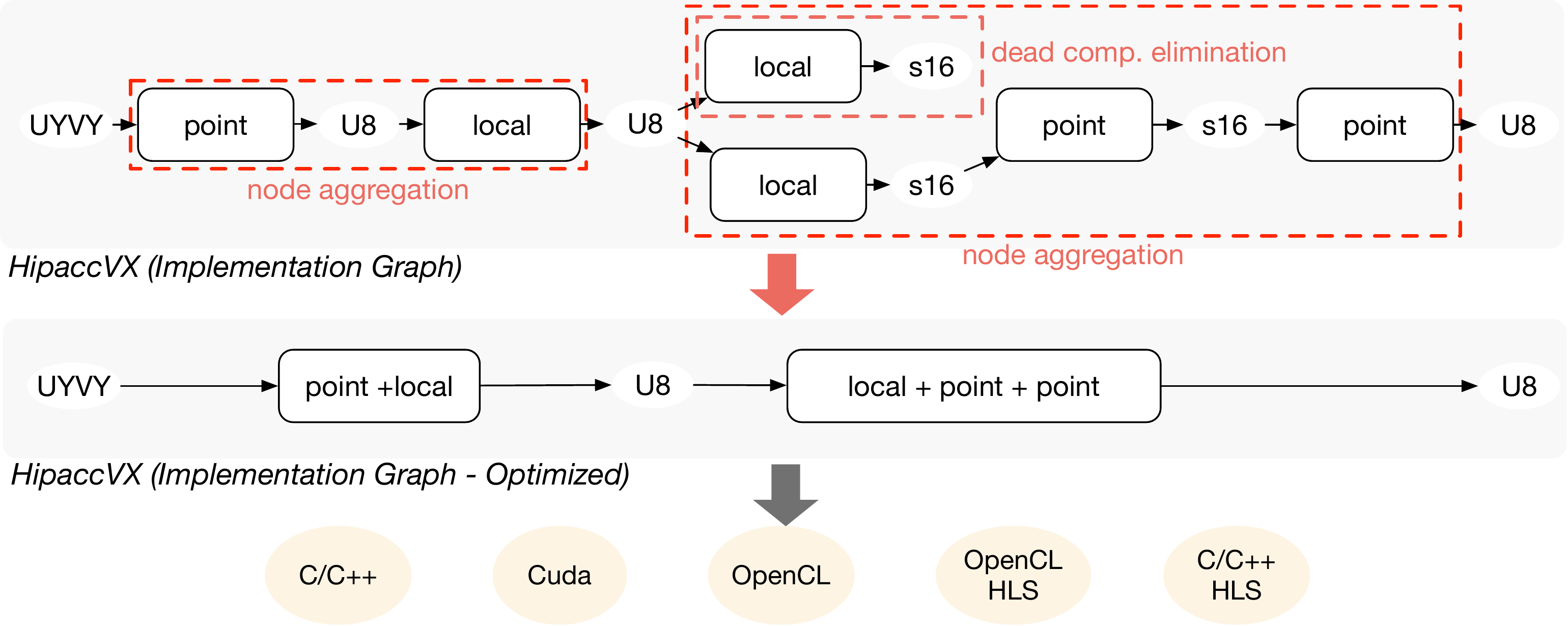}\vspace{\nspace}
  \captionsetup[figure]{width=0.91\linewidth}
\captionof{figure}{
  The application graph in \cref{fig:teaser-appgraph} is implemented by using high-level abstractions called point and local (explained \cref{sec:abstractions}) instead of \OpenVX vision function.
This enables high-performance code generation for various targets when coupled with a DSL compiler and additional optimizations such as dead computation elimination and node aggregation (see \cref{sec:hipaccVX,,sec:domvx-opt}).
  \label{fig:teaser}}
  \vspace{\fspace}
\end{figure*}

\subsection{Combining OpenVX with Image Processing DSLs}\label{sec:combination}

Our solution to the posed challenges in \cref{sec:prob-openvx} is introducing an orthogonal set of so-called \emph{computational abstractions} that enables high-performance implementations for a variety of computing platforms (such as CPUs, GPUs, FPGAs), similar to the \acp{DSL} discussed in \cref{sec:hipacc}.
These abstractions should be used to implement \OpenVX' \ac{CV} functions and, at the same time, be served to the user for the definition of custom nodes.

Assume that the geometric shapes in \cref{fig:customNode} represent the abstractions above.
By implementing both the \OpenVX \ac{CV} functions and the custom node using the basic building block (different geometric shapes in the figure), a consistent graph is constructed for the implementation.
Consequently, the problem of instantiating the user code as a black box is eliminated.
Likewise, assume that all the \ac{CV} functions of the \OpenVX code in \cref{lst:openvx-ex} are implemented by using the computational abstractions called \emph{point} and \emph{local}(explained in \cref{sec:abstractions}).
Then, its application graph (\cref{fig:teaser-appgraph}) transforms into the implementation graph shown in \cref{fig:teaser}.
This implementation graph could be used for target-specific optimizations and code generation similar to the \ac{DSL} compiler approaches for image processing.

In this paper, we implement the \OpenVX standard by the computational abstractions explained in \cref{sec:abstractions}.
We accomplish this task by developing a back end for \OpenVX using \hipacc (as an existing image processing \ac{DSL}) instead of standard programming languages. In this way, we get the best of both worlds (\OpenVX and \ac{DSL} works).
Our approach relies on \OpenVX' industry-standard graph specification and enables \ac{DSL}-based code generation.
The user is offered well-known \ac{CV} functions as well as \ac{DSL} elements (\ie programming constructs, abstractions) for the description of custom nodes.
As a result of this, programmers can write functional descriptions for custom nodes without having concerns about the performance; and, as a consequence, allows writing performance-portable \OpenVX programs for a larger algorithm space.

\begin{figure*}
  \centering
    \subfloat[Point Operator]{\label{fig:pOp}\includegraphics[width=0.26\linewidth]{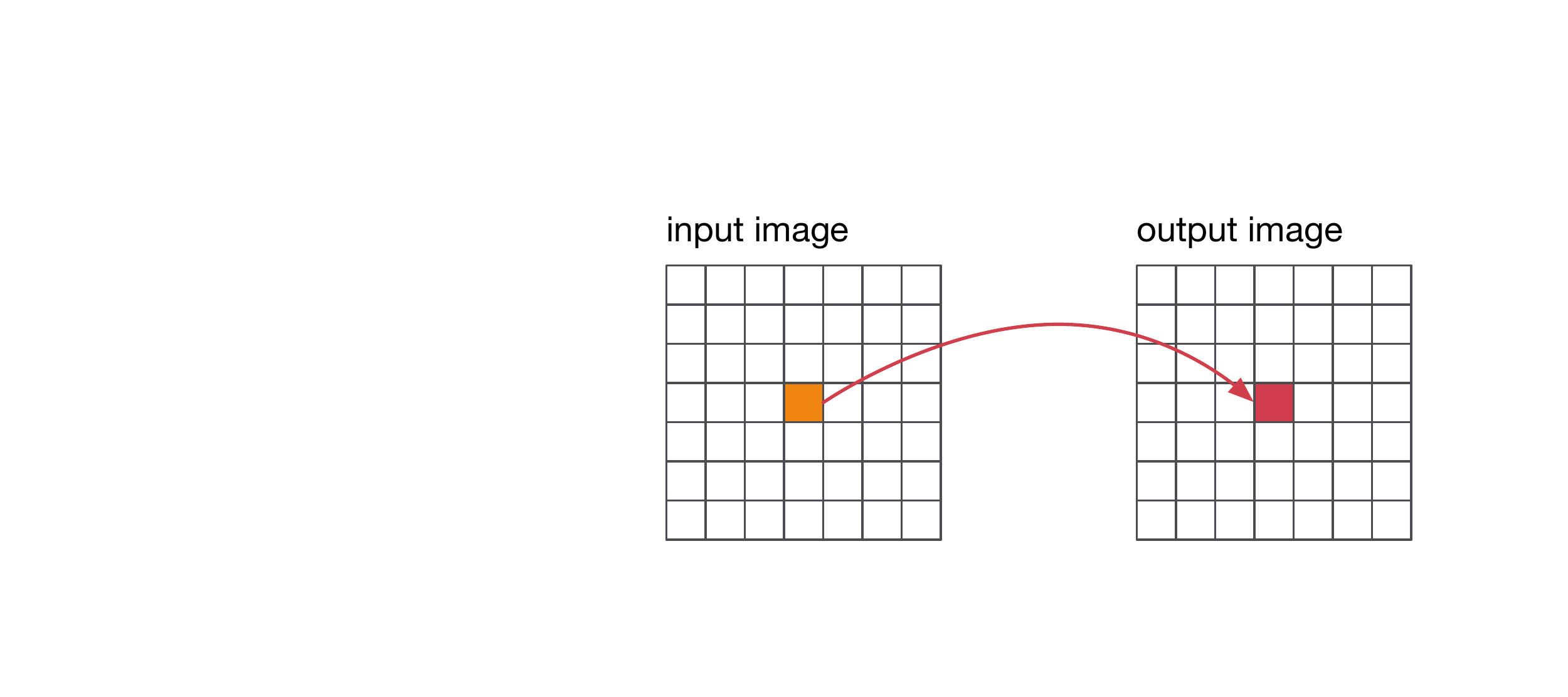}}
    \hspace{2em}
    \subfloat[Local Operator]{\label{fig:lOp}\includegraphics[width=0.26\linewidth]{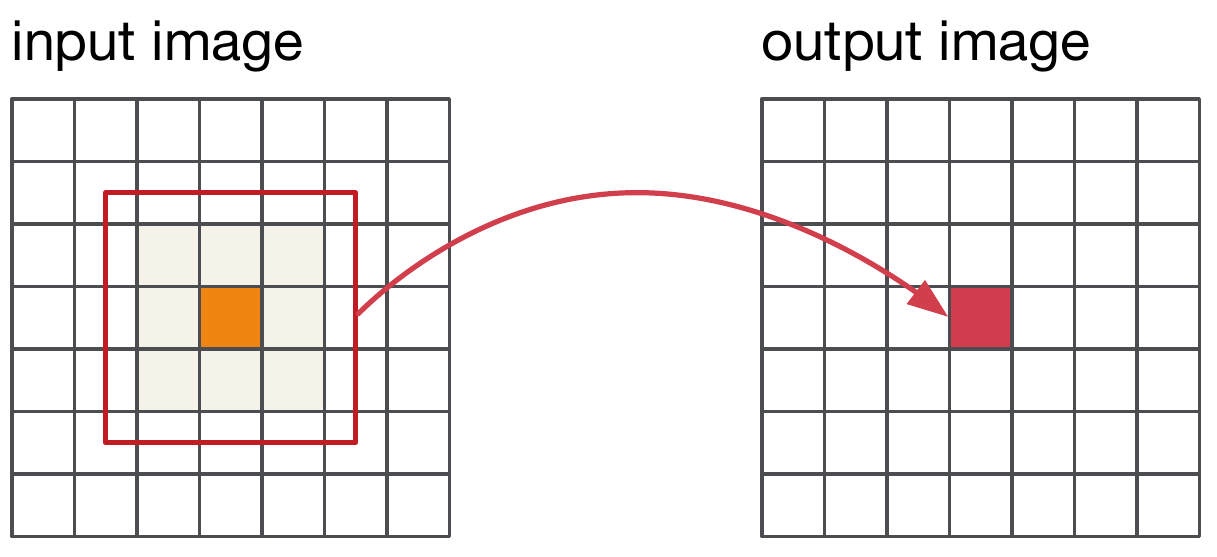}}
    \hspace{2em}
    \subfloat[Global Operator]{\label{fig:gOp}\includegraphics[width=0.26\linewidth]{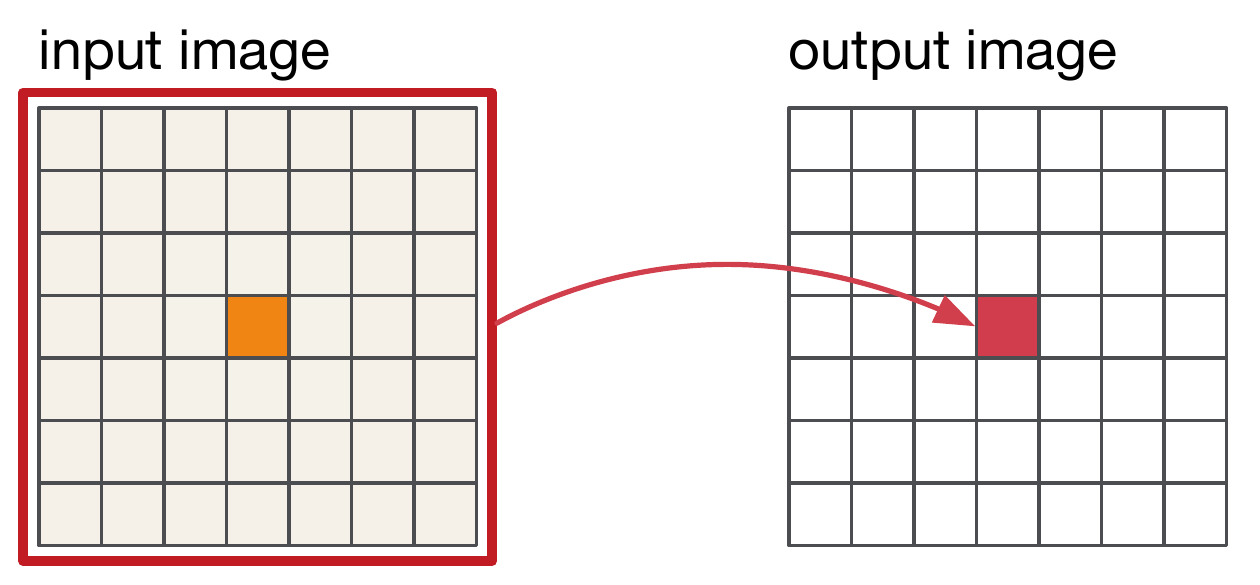}}
  \caption{The considered computational abstractions (listed in \cref{tab:parallel-patterns}) are based on three groups of operators.}
  \label{fig:opTypes}
\end{figure*}

 \begin{table*}[!t]
\ifx\squeeze\empty
    \renewcommand{\arraystretch}{1.5}
\else
    \renewcommand{\arraystretch}{1.7}
\fi
     \newcommand\scalList{ScalarOperation, Select, Remap}
\newcommand\pOpList{AbsDiff, Copy,
                         Add, Subtract, And, Xor, Or, Not, ChannelCombine, ChannelExtract,
                         ColorConvert, ConvertDepth, Magnitude, Phase, Multiply,
                         ScaleImage, Threshold, TensorAdd, TensorSubtract,
                         TensorConvertDepth, TensorMultiply}
     \newcommand\lOpList{NonMaxSuppression, Dilate3x3, Erode3x3,
                         NonLinearFilter, Median3x3, BilateralFilter, Sobel3x3, Box3x3, Convolve, Gaussian3x3, LBP, FastCorners}
     \newcommand\redList{MinMaxLoc, MeanStdDev, Min, Max}
     \newcommand\hisList{Histogram}
     \newcommand\scaList{scale-image}
     \newcommand\pyrList{GaussianPyramid, LaplacianPyramid, LaplacianReconstruct}
     \newcommand\tranList{TensorTranspose, TensorMatrixMultiply}
     \newcommand\warpList{WarpAffine, WarpPerspective}
     \newcommand\checkList{IntegralImage}
\newcommand\habst[1]{#1}
     \newcommand\dabst[1]{\multicolumn{1}{m{8em}}{#1}}
\vspace{\nspace}
    \caption{Categorization of the OpenVX Kernels according to data access patterns
     }
    \label{tab:parallel-patterns}
\centering
      \scalebox{0.8}{
        \label{tab:kernel-classification}
        \begin{tabular}{p{34em}>{\hspace{-45pt}}r<{\hspace{-2pt}}r}
            \toprule\addlinespace[-1mm]
            OpenVX Kernels           & \DomVX Abstractions                    & \habst{Hipacc Abstractions} \\[-1mm]
            \midrule\addlinespace[-1mm]
            {\pOpList, \scalList}    & point                          & \habst{Kernel}    \\[-1mm]
            {\lOpList}               & local                          & \habst{Kernel}    \\[-1mm] {\redList}               & reduce (global)                & \habst{Reduction} \\[-1mm]
            {\hisList}               & histogram (global)             & \habst{Histogram} \\[-1mm]
            {\scaList}               & scale (global)                 & \habst{Interpolation} \\[-1mm]
            {\pyrList}               & pyramid (global)               & \habst{Pyramid} \\[-1mm]
            {\checkList}             & scan (global)                  & \habst{Software}\\[-1mm]
            {\warpList}              & warp (global)                  & \habst{Software} \\[-1mm]
            {\tranList}              & (global) transpose, matrixMult         & \habst{Software} \\[-1mm]
            HarrisCorners            & point + local + \emph{custom}  & \habst{Kernel, Software}  \\[-1mm]
            EqualizeHist             & histogram + point              & \habst{Kernel, Histogram} \\[-1mm]
            OpticalFlowPyrLK         & point + local + pyramid + \emph{custom}& \habst{Kernel, Pyramid, Software}\\[-1mm]
            HOGCells                 & \emph{custom} + local + histogram      & \habst{Kernel, Software} \\[-1mm]
            CannyEdge                & point + local + \emph{custom}          & \habst{Kernel, Software} \\[-1mm]
            \bottomrule
         \end{tabular}
      }

\vspace{\fspace}
\end{table*}

\section{Computational Abstractions}\label{sec:abstractions}

We have analyzed \OpenVX' \ac{CV} functions and categorized them into the computational abstractions summarized in \cref{tab:parallel-patterns}.
The categorization is mainly based on three groups of operators: \begin{enumerate*} [label=(\itshape\roman*\upshape)]
\item \emph{point operators} that compute an output from one input pixel,
\item \emph{local operators} depend on neighbor pixels over a certain region, and
\item \emph{global operators} where the output might depend on the whole input image, (presented in \cref{fig:opTypes}).\end{enumerate*}
We have identified the following patterns for the global operators:
\begin{enumerate*} [label=(\itshape\alph*\upshape)]
\item \emph{reduction}: traverses an input image to compute one output (\eg max, mean),
\item \emph{histogram}: categorizes (maps) input pixels to bins according to a binning (reduce) function,
\item \emph{scaling}: downsizes or expands input images by interpolation,
\item \emph{scan}: each output pixel depends on the previous output pixel.
\end{enumerate*}
\emph{Warp}, \emph{transpose}, and \emph{matrix multiplication} are denoted as global operator blocks.

Through the introduction of the node-internal computational abstractions, our approach enables additional optimizations that manipulate the computation (see \cref{sec:domvx-opt,,sec:hipaccVX}).
This is also illustrated in \cref{fig:teaser}, where redundant computations are eliminated, and nodes are aggregated for better exploitation of locality.
Memory access patterns of our abstractions entail system-level optimization strategies motivated by the \OpenVX standard, such as image tiling~\cite{tagliavini2016optimizing} and hardware-software partitioning~\cite{taheri2018acceleration}.
An abstraction-based implementation allows expressing aggregated computations as part of the reconstructed graph.
In this way, an implementation graph, as well as an application graph can be expressed using the same graph structure.
Furthermore, using the proposed set of abstractions reduces code duplication compared to typical approaches, where the libraries are implemented using hand-written \ac{CV} functions.
For instance, 36 of \OpenVX' \ac{CV} functions can be implemented solely with the description of point and local operators as shown in \cref{tab:parallel-patterns}; that is, a few highly optimized building blocks for a single target platform (\eg GPU) can be reused.

 \section{The \DomVX Framework}\label{sec:implementation}

In this paper, we developed a framework, called \DomVX, which is a DSL-based implementation of \OpenVX.
We extended \OpenVX specification by \hipacc code interoperability (see \cref{sec:dsl-backend}) such that
programmers are allowed to register \hipacc kernels as custom nodes to \OpenVX programs.
The \DomVX framework consists of an \OpenVX graph implementation and optimization routines that verify and optimize input \OpenVX applications (see \cref{sec:domvx-opt}).
Ultimately, it generates a device-specific code for the target platform using \hipacc's code generation.
The tool flow is presented in \cref{fig:hipaccVX_overview}.

\subsection{DSL Back End and User-Defined Kernels}\label{sec:dsl-backend}

\OpenVX mandates the verification of parameters and the relationship between input and output and parameters as presented in \cref{lst:openvx-custom}.
There, first, a user kernel and all of its parameters should be defined (\crefrange{ll:vx-extension}{ll:vx-addparam2}).
Then, a custom node should be created by \lst[language=OpenVX]{vxCreateGenericNode} (\cref{ll:vx-node}) after the user kernel is finalized by a \lst[language=OpenVX]{vxFinalizeKernel} call (\cref{ll:vx-kernel-finalize}).
The kernel parameter types are defined, and the node parameters are set by
\lst[language=OpenVX]{vxAddParameterToKernel} (\crefrange{ll:vx-addparam0}{ll:vx-addparam2}) and \lst[language=OpenVX]{vxSetParameterByIndex} (\crefrange{ll:vx-setparam0}{ll:vx-setparam2}), respectively.

We extended \OpenVX by \lst[language=OpenVX]{vxHipaccKernel} function (\cref{ll:vx-extension}) to instantiate a \hipacc kernel as an \OpenVX kernel.
The \hipacc kernels should be written in a separate file and added as a generic node according to the \OpenVX standard~\cite{OpenVX13}.
Programmers do not have to describe the dependency between \hipacc kernels as in \cref{lst:hipacc-host}, instead, they write a regular \OpenVX program to describe an application graph.
This sustains the custom node definition procedure of \OpenVX.
Ultimately, the \DomVX framework verifies and optimizes a given \OpenVX application, generates the corresponding \hipacc code, and employs \hipacc for device-specific code generation.

\begin{lstlisting}[
    float =!t,
    basicstyle=\scalefont{.71}\tt,
    language=OpenVX,
    frame=tb,
    caption={DSL code interoperability extension (only \cref{ll:vx-extension}).},
    label={lst:openvx-custom},
    ]
vx_node vxGaussian3x3Node(vx_graph graph,
                          vx_image arr,
                          vx_image out) {

  // Extension: An OpenVX kernel from a Hipacc kernel
  vx_kernel cstmk = vxHipaccKernel("gaussian3x3.cpp");$\label{ll:vx-extension}$

  /*** The code below is the standard OpenVX API ***/
  // Create vx_matrix for mask
  const float coeffs[3][3] = /* ... */;
  vx_matrix mask = vxCreateMatrix(context,
                                  VX_TYPE_FLOAT32,
                                  3, 3);
  vxCopyMatrix(mask, (void*)coeffs, VX_WRITE_ONLY,
               VX_MEMORY_TYPE_HOST);

  // Set input/output parameters for a kernel
  vxAddParameterToKernel(cstmk, 0, VX_OUTPUT,
                         VX_TYPE_IMAGE,
                         VX_PARAMETER_STATE_REQUIRED);$\label{ll:vx-addparam0}$
  vxAddParameterToKernel(cstmk, 1, VX_INPUT,
                         VX_TYPE_IMAGE,
                         VX_PARAMETER_STATE_REQUIRED);
  vxAddParameterToKernel(cstmk, 2, VX_INPUT,
                         VX_TYPE_MATRIX,
                         VX_PARAMETER_STATE_REQUIRED);$\label{ll:vx-addparam2}$
  vxFinalizeKernel(cstmk);$\label{ll:vx-kernel-finalize}$

  // Create generic node
  vx_node node = vxCreateGenericNode(graph, cstm_k);$\label{ll:vx-node}$
  vxSetParameterByIndex(node, 0, (vx_reference) out);$\label{ll:vx-setparam0}$
  vxSetParameterByIndex(node, 1, (vx_reference) arr);
  vxSetParameterByIndex(node, 2, (vx_reference) mask);$\label{ll:vx-setparam2}$

  return node;
}
\end{lstlisting}

\OpenVX' \ac{CV} functions are implemented as a library by using our extension for \hipacc code instantiation.
For instance, the \DomVX implementation of the \lst[language=OpenVX]{vxGaussian3x3Node} API is shown in \cref{lst:openvx-custom}.
Users can simply use these \ac{CV} functions as in \cref{lst:openvx-ex}.
A minority of \OpenVX functions are implemented as OpenCV kernels since they cannot be fully described in \hipacc.
These are listed in \cref{tab:parallel-patterns} with a \emph{Software} label instead of a \hipacc abstraction type.
As future work, we can extend \hipacc to support these functions.

\subsubsection{Optimizations Based on Code Generation}\label{sec:hipaccVX}

We inherited many device-specific optimization techniques by implementing a \hipacc back end for \OpenVX.
\hipacc internally applies several optimizations for the code generation from its DSL abstractions.
These include memory padding, constant propagation, utilization of textures, loop
unrolling, kernel fusion, thread-coarsening, implicit use of unified CPU/GPU memory, and the integration with CUDA Graph~\cite{membarth2016hipacc,reiche2017auto,boFusion,boCudaGraph}.
At the same time, \hipacc targets Intel and Xilinx \acp{FPGA} using their \ac{HLS} tools.
There, an input application is implemented through application circuits derived from the \ac{DSL} abstractions and optimized by hardware techniques such as pipelining and loop coarsening~\cite{hipaccIccad,ozkan2016fpga,ASAP17}.

\subsection{\OpenVX Graph and System-Level Optimizations}\label{sec:domvx-opt}

As mentioned before, an \OpenVX application is represented by a \ac{DAG} $G_{app}=(V,E)$, where $V$ is a set of vertices, and $E$ is a set of edges $E \subseteq V \times V$ denoting data dependencies between nodes.
The set of vertices $V$ can further be divided into two disjoint sets $D$ and $N$ ($V = D \cup N$, $D \cap N = \emptyset$) denoting data objects and \ac{CV} functions, respectively.

Both data (\ie \lst{Image}, \lst{Scalar}, \lst{Array}) and node (\ie \ac{CV} functions) objects are implemented as \cpp classes that inherit the \OpenVX \lst{Object} class.
Vertices $v \in V$ of our \OpenVX graph implementation consist of \OpenVX \lst{Object} pointers.
The verification phase first checks if an application graph $G_{app}$ (derived from the user code, see, \eg \cref{lst:openvx-ex}) does not contain any cycles.
Then, it verifies that the description is a bipartite graph, \ie $\forall (v,w) \in E :\ v \in D \land w \in N \lor v \in N \land w \in D$.
Finally, the verification phase applies the following optimizations:
\color{black}

\subsubsection{Reduction of Data Transfers}\label{sec:dataflow}

Data nodes of an application graph that are not virtual must be accessible to the host, while the intermediate (virtual) points of a computation should be stored in the device memory.
We distinguish these two data node types by the set of non-virtual data nodes $D_{nv}$ and the set of virtual data nodes $D_{v}$, where $D = D_{nv} \cup D_{v}$, $D_{nv} \cap D_{v} = \emptyset$.
\DomVX keeps this information in its graph implementation and determines the subgraphs between non-virtual data nodes, which can be kept in the device memory.
In this way, data transfers between host and device are avoided.

\subsubsection{Elimination of Dead Computations}\label{sec:domvx-opt-comp}
An application graph may consist of nodes that do not affect the results.
Inefficient user code or other compiler transformations might cause such dead code.
A less apparent reason could be the usage of \OpenVX compound \ac{CV} functions for smaller tasks.
Consider \lst[language=OpenVX]{Sobel3x3} as an example, which computes two images, one for the horizontal and one for the vertical derivative of a given image.
As the \OpenVX API does not offer these algorithms separately, programmers have to call \lst[language=OpenVX]{Sobel3x3}, even when  they are only interested in one of the two resulting images.
Our implementation is based on abstractions and allows a better analysis of the computation compared to \OpenVX' \ac{CV} functions, \ie the Sobel API is implemented by two parallel local operators as shown in \cref{fig:teaser}.
\DomVX optimizes a given application graph using the procedure described in \cref{alg:deathCompElimination}.
Conventional compilers do not analyze this redundancy if utilizing the host/device execution paradigm (\eg OpenCL, CUDA); that means, when \OpenVX kernels are offloaded to an accelerator device, and device kernels are executed by the host according to the application dependency (see \cref{sec:res-optimizations}).

\Cref{alg:deathCompElimination} assumes that the non-virtual data nodes whose input and output degrees are zero must be the inputs ($D_{in}$) and the results ($D_{out}$) of an application, respectively.
Other non-virtual data nodes could be input, output, or intermediate points of an application depending on the number of connected virtual data nodes.
These are initialized in \cref{algOptDeath:beg}.
Then, all of the nodes in the same component between the node $v_{start}$ and the set $V_{in}$ are traversed via the \emph{depth-first visit} function (\cref{algOptDeath:dfs}) and marked as alive (Lines~\ref{algOptDeath:beg}~to~\ref{algOptDeath:end}).
Finally, in \cref{algOptDeath:filt}, a filtered view of an application graph is created from the set of alive nodes.

The complexity of the functions transpose (\cref{algOptDeath:trans}) and depth-first visit (\cref{algOptDeath:dfs}) are $\mathcal{O}(|V| + |E|)$ and $\mathcal{O}(|E|)$, correspondingly.
The filter graph function (\cref{algOptDeath:filt}) is only an adaptor that requires no change in the application graph~\cite{siek2002boost}.
In the worst case, the graph has $|V|-2$ output data nodes.
That is, the complexity of \cref{alg:deathCompElimination} becomes $\mathcal{O}(|V|^{2} + |E|)$ in time and $\mathcal{O}(|V| + |E|)$ in space.

\begin{algorithm}[!t]
  \scriptsize
  \SetAlCapFnt{\footnotesize}
  \SetAlCapNameFnt{\footnotesize}

  \DontPrintSemicolon
  \SetKwProg{Func}{function}{}{end}
	\SetKwInOut{Input}{input}
  \SetKwInOut{Output}{output}

  \SetKwFunction{inedge}{deg$^{-}$}
  \SetKwFunction{outedge}{deg$^{+}$}
  \SetKwFunction{main}{eliminate\_death\_nodes}
  \SetKwFunction{cycles}{if\_cycle\_exist}
  \SetKwFunction{init}{init\_root\_and\_child\_nodes}
  \SetKwFunction{transpose}{transpose\_graph}
  \SetKwFunction{filter}{filter\_graph}
  \SetKwFunction{alive}{mark\_as\_alive}
  \SetKwFunction{visit}{depth-first\_visit}

  \SetKwData{NodesOrdered}{$V_{ordered}$}
  \SetKwData{NodesData}{$D_{nv}$}
  \SetKwData{NodesDin}{$D_{in}$}
  \SetKwData{NodesDout}{$D_{out}$}
  \SetKwData{NodesVisit}{$V_{v}$}
  \SetKwData{NodesAlive}{$V_{alive}$}
  \SetKwData{vert}{$v$}
  \SetKwData{start}{$v_{start}$}

  \SetKwData{Gemp}{$G_{\emptyset}$}
  \SetKwData{Gapp}{$G_{app}$}
  \SetKwData{Gtran}{$G_{trans}$}
  \SetKwData{Gfilt}{$G_{filt}$}

  \Input{\Gapp\ --\ application graph \\
				 \NodesData\ --\ set of are non-virtual data nodes}
	\Output{\Gfilt\ --\ optimized application graph}

  \Func{\main{\Gapp, \NodesData}} {
			\tcc{Find candidate non-virtual roots and leaves}
  		$\NodesDin \gets \emptyset$, $\NodesDout \gets \emptyset$\;  \label{algOptDeath:beg}
	   	\ForAll{\vert $\in$ \NodesData} {
					\If {\inedge(\vert) = 0} {
							\NodesDin  $\gets$ \NodesDin  $\cup$ \vert \tcp*{input non-virtual data nodes}
					}
					\ElseIf { \outedge(\vert) = 0} {
							\NodesDout $\gets$ \NodesDout $\cup$ \vert \tcp*{out non-virtual data nodes}
					}
					\Else { \NodesDin  $\gets$ \NodesDin  $\cup$ \vert \;
									\NodesDout $\gets$ \NodesDout $\cup$ \vert }
    	}                          																	 \label{algOptDeath:end}

			\tcc{Mark the nodes between roots and leaves as alive}
			\Gtran $\gets$ \transpose(\Gapp)\;													 \label{algOptDeath:trans}
$\NodesAlive \gets \emptyset$\;                              \label{algOptDeath:opt1}
	   	\ForAll{\start $\in$ \NodesDout} {
					\NodesVisit  $\gets$ \visit(\start, \NodesDin, \Gtran)\; \label{algOptDeath:dfs}
          \NodesAlive  $\gets$ \NodesAlive $\cup$ \NodesVisit
			}                                                            \label{algOptDeath:end}
\tcc{Filter, keep only the alive nodes and their edges}
			\Gfilt$\gets$ \filter(\Gapp, KEEP\_EDGES, \NodesAlive)\;      \label{algOptDeath:filt}

			\Return $\Gfilt$\;
  }
  \caption{Graph Analysis for Dead Computation Elimination}
  \label{alg:deathCompElimination}
\end{algorithm}
  \section{Evaluation and Results}\label{sec:results}
We present results for a Xilinx Zynq ZYNQ-zc706 \ac{FPGA} using Xilinx Vivado HLS 2019.1 and an Nvidia GeForce GTX~680 with CUDA driver 10.0.
We evaluate the following applications:
As image smoothers, we consider a Gaussian blur (\emph{Gauss}) and a \emph{Laplacian} filter  with a $5\times5$ and $3\times3$ local node, respectively.
The filter chain (\emph{FChain}) is an image pre-processing algorithm consisting of three convolution (local) nodes.
The \emph{SobelX} determines the horizontal derivative of an input image using the \OpenVX \lst{vxSobel} function.
The edge detector in \cref{fig:teaser-appgraph} (\emph{EdgFig\ref{fig:teaser-appgraph}}) finds horizontal edges in an input image, while \emph{Sobel} computes both horizontal and vertical edges using three \ac{CV} nodes.
The \emph{Unsharp} filter sharpens the edges of an input image using one Gauss node and three point operator nodes.
Both \emph{Harris} and \emph{Tomasi} detect corners of a given image using $13$ ($4$ local + $9$ point) and $14$ ($4$ local + $10$ point) \ac{CV} nodes, respectively.
These applications are representative to show the optimization techniques discussed in this paper. The performance of a simple  \ac{CV} application (\eg \emph{Gauss}) solely depends on the quality of code generation, while graph-based optimizations can further optimize the performance of more complex applications (\eg \emph{Tomasi}).
\emph{Laplacian} uses the \OpenVX' custom convolution API and \emph{EdgFig2} consists of redundant kernels.

\subsection{Acceleration of User-Defined Nodes}\label{sec:res-cust}
\begin{figure}
    \vspace{\nspace}
    \def\datafile{data/custom_node_acc_gpu.dat}
    \raggedleft
    \scalebox{1}{\includegraphics{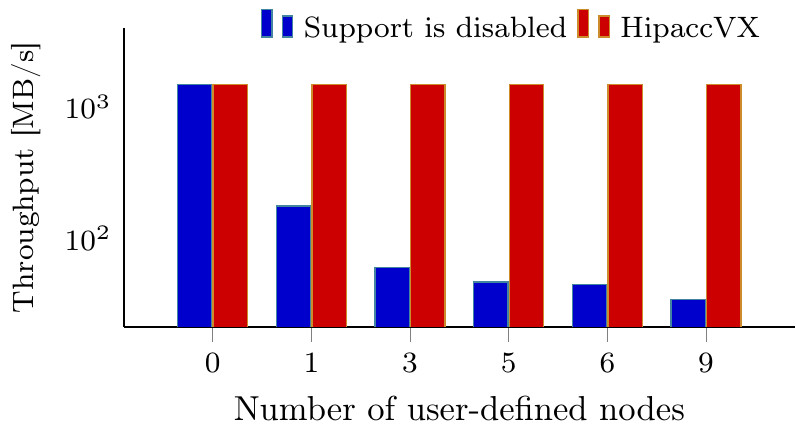}}
    \vspace{\fspace}
    \caption{
      Throughput for different versions  of the same corner detection application (consisting of 9 kernels) on the Nvidia GTX680 (higher is better).
      The blue bars denote an increasing number of \ac{CV} functions implemented as user-defined nodes using \cpp.
      In \OpenVX, these user-defined functions have to be executed on the host CPU, which leads to a performance degradation; whereas, \DomVX accelerates all user-defined nodes on the GPU.
    }
    \vspace{2.1\nspace}
  \label{fig:cust-node-gpu}
\end{figure}

User-defined nodes can be accelerated on a target platform (\eg GPU accelerator) when they are expressed with \DomVX' abstractions (see \cref{sec:dsl-backend}).
A \cpp implementation of these custom nodes results in executing them on the host device.
This is illustrated in \cref{fig:cust-node-gpu} for a corner detection algorithm that consists of nine kernels. The {CPU} codes for these custom nodes are also acquired using Hipacc.
As can be seen in \cref{fig:cust-node-gpu}, \DomVX provides the same performance invariant to the number of user-defined nodes, whereas using the \OpenVX API decreases the throughput severely since each user-defined node has to be executed on the host CPU.

\subsection{System-Level Optimizations based on \OpenVX Graph}\label{sec:res-optimizations}

\begin{figure}
    \hspace{-2em}
    \subfloat[Xilinx Zynq FPGA]{
\scalebox{0.8}{\includegraphics{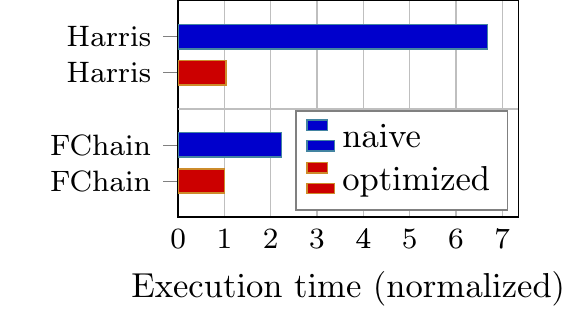}}
    		\label{fig:reduce-data-trans-fpga}
    }
    \subfloat[Nvidia GTX~680 GPU]{
    \hspace{-2em}
\scalebox{0.8}{\includegraphics{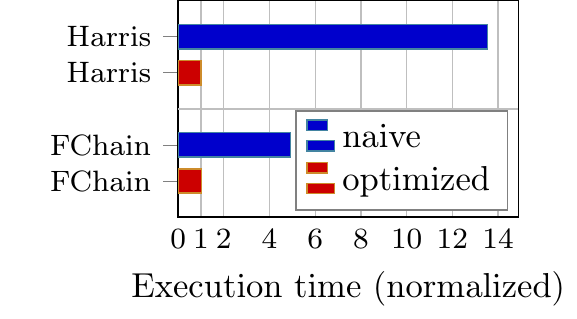}}
    		\label{fig:reduce-data-trans-gpu}
		}
    \vspace{\subfspace}
    \caption{
        Normalized execution time (lower is better) for $1024\times1024$ images.
 				\DomVX eliminates redundant transfers by analyzing \OpenVX' graph-based application code.
    }
    \label{fig:reduce-data-trans}
    \vspace{-1em}
\end{figure}

\paragraph{Reduction of Data Transfers}
\DomVX eliminates the data transfers between the execution of subsequent functions on a target accelerator device, as explained in \cref{sec:dataflow}.
This is disabled for \emph{naive} implementations.
The improvements for the two applications are shown in \cref{fig:reduce-data-trans}.
\DomVX' throughput optimizations reach a speedup of 13.5.

\begin{figure}
    \hspace{-2em}
    \subfloat[Xilinx Zynq FPGA]{
    		\scalebox{0.8}{\includegraphics{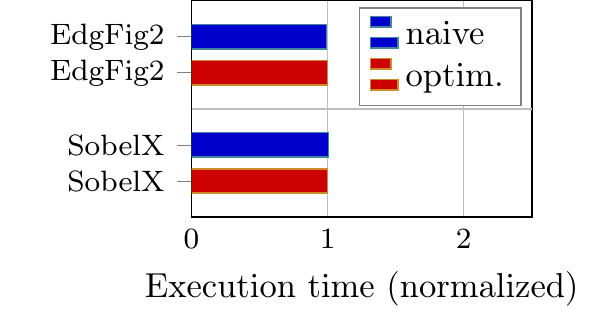}}
    		\label{fig:dead-node-fpga}
    }
    \subfloat[Nvidia GTX~680 GPU]{
      \hspace{-2em}
      \scalebox{0.8}{\includegraphics{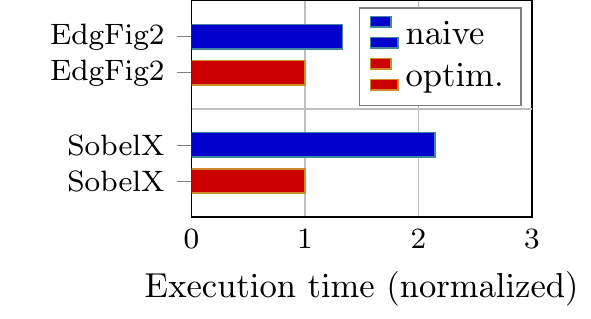}}
        \label{fig:dead-node-gpu}
		}
    \vspace{\subfspace}
    \caption{
        Normalized execution time (lower is better) for $1024\times1024$ images.
}
    \label{fig:dead-node}
    \vspace{1.1\nspace}
\end{figure}

\begin{figure}
    \def\datafile{data/dead_node_fpga_area.dat}
    \begin{center}
        \scalebox{0.9}{\includegraphics{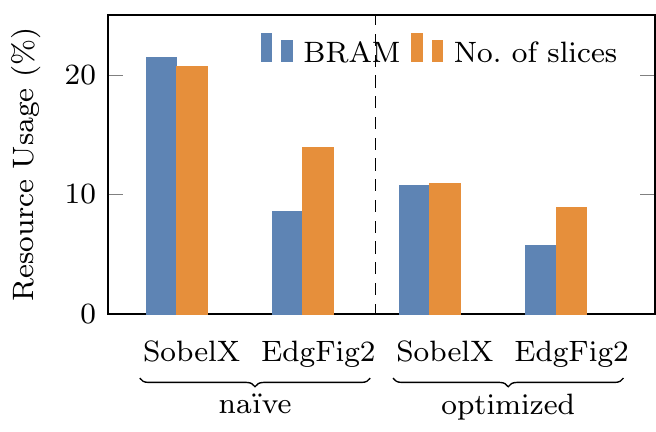}}
    \end{center}
    \vspace{-1.1em}
    \caption{
      \ac{PPnR} results for the Xilinx Zynq FPGA.
      Elimination of dead computation reduces the area, significantly.
    }
    \vspace{1.1\nspace}
  \label{fig:dead-node-fpga-area}
\end{figure}

\paragraph{Elimination of Dead Computation}
\DomVX eliminates the computations that do not affect the results of an application (see \cref{sec:domvx-opt-comp}).
This is illustrated in \cref{fig:dead-node}.
\DomVX improves the throughput by a factor of 2.1 on the GTX~680.
The throughput improvement for the Zynq FPGA is only slightly better since the applications fit into the target device; thus, run in parallel.
Yet, \DomVX' FPGA implementation for the same application reduces the number of FPGA resources (elementary programmable logic blocks called \emph{slices} and on-chip block RAMs, short \emph{BRAMs}) significantly (around 50\% for SobelX) on the Zynq (see Fig.~\ref{fig:dead-node-fpga-area}).

\subsection{Evaluation of the Performance}\label{sec:res-comp}

\begin{figure}
  \def\leftsep{\hspace{0em}}\begin{center}
   \scalebox{1}{\includegraphics{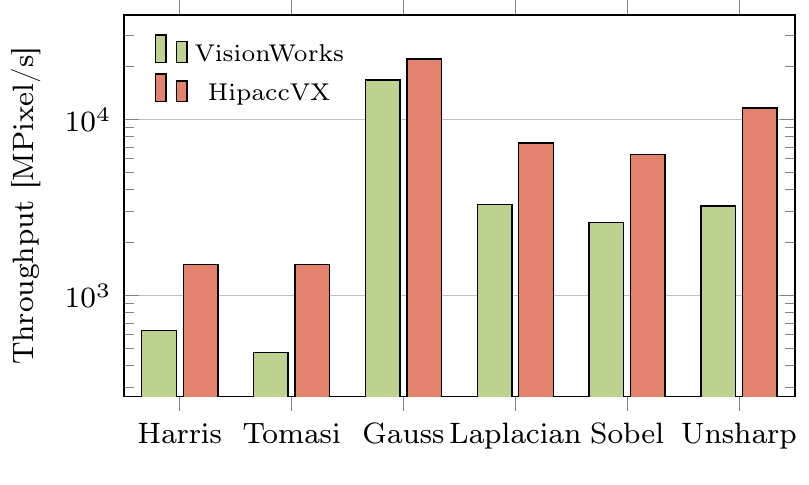}}
 \end{center}
\vspace{\fspace}
   \caption{
			Comparison of Nvidia VisionWorks~v1.6 and \DomVX on the Nvidia GTX~680.
		  Image sizes are $2048\times2048$.
		}
		\vspace{\nspace}
  \label{fig:compGPU}
\end{figure}

In \cref{fig:compGPU}, we compare \DomVX with the VisionWorks~(v1.6) provided by Nvidia, which provides an optimized commercial implementation of \OpenVX.
\DomVX, as well as typical library implementations, exploit the graph-based \OpenVX API to apply system-level optimizations~\cite{rainey2014addressing}, such as reduction of data transfers (see \cref{sec:domvx-opt}).
Additionally, \DomVX generates code that is specific to target GPU architectures and applies optimizations such as constant propagation, thread coarsening, \ac{MPMD}~\cite{membarth2016hipacc}.
As shown in \cref{fig:compGPU},
\DomVX can generate implementations that provide higher throughput than VisionWorks.
Here, the speedups for applications that are composed of multiple kernels (Harris, Tomasi, Sobel, Unsharp) are higher than the ones solely consisting of one OpenVX~CV function (Gauss and Laplacian).
This performance boost is, to a large extent, due to the locality optimization achieved by fusing consecutive kernels at the compiler level~\cite{boFusion}.
This requires code rewriting and the resource analysis of the target GPU architectures.

There was no publicly available \ac{FPGA} implementation of OpenVX at the time this paper was written.
Therefore, in \cref{tab:dsl-comp-hls}, we compare \DomVX with Halide-HLS~\cite{pu2017programming}, which is a state-of-the-art DSL targeting Xilinx FPGAs.
As can be seen, \DomVX uses fewer resources and achieves a higher throughput for the benchmark applications.
\begin{table}[!ht]
    \vspace{2\subfspace}
    \caption{
    \ac{PPnR} results for the Xilinx Zynq for images of $1020\times1020$ and $T_\mathit{target}$ = 5 ns (corresponds to $f_\mathit{target}$ = 200 MHz).
    }
    \vspace{\fspace}
  	\vspace{2\spaceoffset}
    \centering
    \resizebox{\columnwidth}{!}{
        \begin{tabular}{lllrrrr}
App                     & v                  &              &   BRAM &  SLICE & DSP & Latency [cyc.]  \\
\toprule
\multirow{4}{*}{Gauss}  & \multirow{2}{*}{1} & \DomVX       &      8 &    473 & 16  & 1044500 \\
                        &                    & Halide-HLS   &      8 &   1823 & 50  & 1052673 \\
\cmidrule{3-7}
                        & \multirow{2}{*}{4} & \DomVX       &     16 &   1519 & 64   &  261649 \\
                        &                    & Halide-HLS   &     16 &   4112 &180   &  266241 \\
\midrule
\multirow{4}{*}{Harris} & \multirow{2}{*}{1} & \DomVX       &     20 &   1457 & 34   & 1042466 \\
                        &                    & Halide-HLS   &     16 &   2688 & 35   & 1052673 \\
\cmidrule{3-7}
                        & \multirow{2}{*}{2} & \DomVX       &     20 &   2326 & 68   &  521756 \\
                        &                    & Halide-HLS   &     16 &   4011 & 70   &  528385 \\
\bottomrule
        \end{tabular}
    }
    \label{tab:dsl-comp-hls}
		\vspace{0.9\fspace}
\end{table}
\DomVX transforms a given \OpenVX application into a streaming pipeline by replacing virtual images with FIFO semantics.
Thereby, it uses an internal representation in \ac{SSA} form.
Furthermore, it replicates the innermost kernel to achieve higher parallelism for a given factor $v$.
For practical purposes, we present results only for Xilinx technology.
Prior work~\cite{ozkan2016fpga,hipaccIccad} shows that \hipacc can achieve a performance similar to handwritten examples provided by Intel for image processing.
This also indicates that the memory abstractions given in~\cref{tab:parallel-patterns} are suitable to generate optimized code for HLS tools.

\begin{figure}[t]
  \def\datafile{data/performance_portability.data}
  \begin{center}
  \hspace{1.5em}
  \scalebox{0.98}{\includegraphics{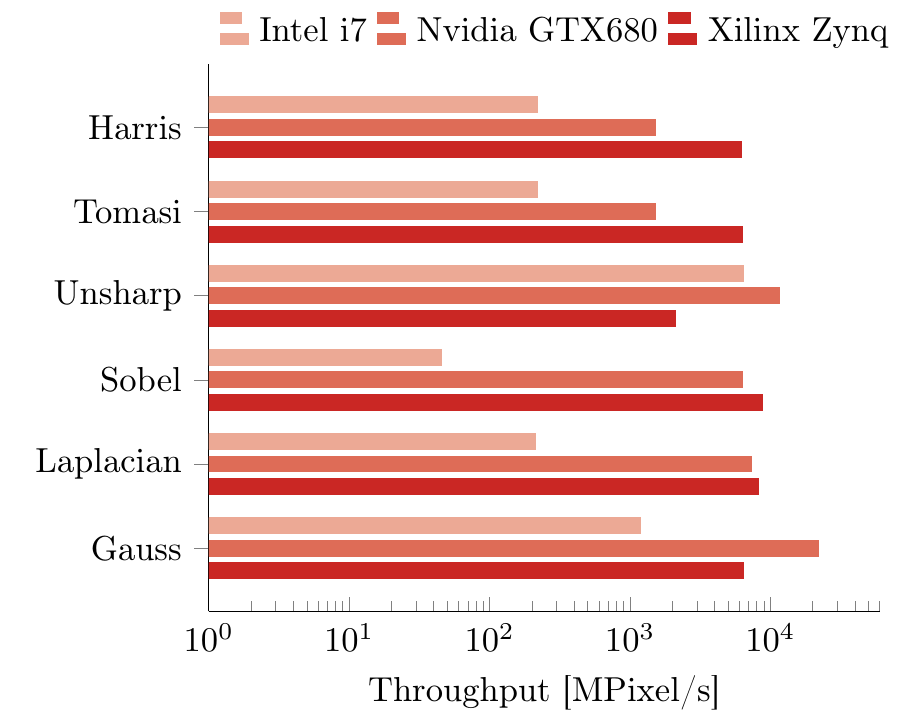}}
  \end{center}
\vspace{\fspace}
  \caption{Comparison of throughput for the Nvidia GTX680, Xilinx Zynq, and Intel i7-4790 CPU.
  The same OpenVX application code is used to generate different accelerator implementations.
  The \DomVX framework allows for both code and performance portability by generating optimized implementations for a diverse range of accelerators.
}
  \label{fig:compPortability}
  \vspace{\nspace}
\end{figure}

\cref{fig:compPortability} compares the throughputs that were achieved from the same \OpenVX application code for different accelerators.
Here, we generated OpenCL, CUDA, and Vivado HLS (\cpp) code to implement a given application on an Intel i7-4790 CPU, an Nvidia GTX680 GPU, and a Xilinx Zynq FPGA, respectively.
GPUs and FPGAs can exploit data-level parallelism by processing a significantly higher number of operations in parallel compared to CPUs.
This makes them very suitable for computer vision applications.
Modern GPUs operate on a higher clock frequency compared to existing FPGAs, therefore they could provide higher throughput for the abundantly parallel applications.
This is the case for Gauss and Unsharp.
Whereas, FPGAs can exploit temporal locality by using \emph{pipelining} and eliminate unnecessary data transfers to global memory between consecutive kernels.
Therefore, all the FPGA implementations in \cref{fig:compPortability} achieve a similar throughput.

\section{Conclusion}\label{sec:conclusion}

In this paper, we presented a set of computational abstractions that are used for expressing \OpenVX' \ac{CV} functions as well as user-defined kernels.
This enables the execution of user nodes on a target accelerator similar to the \ac{CV} functions and additional optimizations that improve the performance.
We presented \DomVX, an implementation for \OpenVX using the proposed abstractions to generate code for \acp{GPU}, CPUs, and \acp{FPGA}.

\bibliographystyle{spmpsci}
\bibliography{references}

\end{document}